\documentclass[letterpaper, 10 pt, conference]{ieeeconf}  

\IEEEoverridecommandlockouts                              

\usepackage{amsmath,amssymb,amsfonts}
\usepackage{algorithmic}
\usepackage{graphicx}
\usepackage{textcomp}
\usepackage[table]{xcolor}
\usepackage{url}
\usepackage{hyperref}
\hypersetup{
    colorlinks=true,
    linkcolor=blue,
    citecolor=black,
    filecolor=magenta,      
    urlcolor=blue,
    pdftitle={Overleaf Example},
    pdfpagemode=FullScreen,
    }

\usepackage[font=small]{caption}

\usepackage[linesnumbered,ruled,vlined]{algorithm2e}
\DeclareMathOperator*{\argmin}{argmin}
\usepackage{array}
\newcolumntype{P}[1]{>{\centering\arraybackslash}p{#1}}
\usepackage{bm}

\usepackage{lipsum} 
\let\labelindent\relax
\usepackage{enumitem}

\title{\LARGE \bf
Object Permanence Filter for Robust Tracking with Interactive Robots
}
\author{Shaoting Peng$^{1}$, Margaret X. Wang$^{2}$, Julie A. Shah$^{2}$ and Nadia Figueroa$^{1}$
\thanks{$^*$Corresponding author: \texttt{pengsht@seas.upenn.edu} }%
\thanks{$^{1}$University of Pennsylvania, Philadelphia, USA}%
\thanks{$^{2}$Massachusetts Institute of Technology, Cambridge, USA}%
}

\begin{document}
\maketitle
\thispagestyle{empty}
\pagestyle{empty}
\begin{abstract}
Object permanence, which refers to the concept that objects continue to exist even when they are no longer perceivable through the senses, is a crucial aspect of human cognitive development. In this work, we seek to incorporate this understanding into interactive robots by proposing a set of assumptions and rules to represent object permanence in multi-object, multi-agent interactive scenarios. We integrate these rules into the particle filter, resulting in the Object Permanence Filter (OPF). For multi-object scenarios, we propose an ensemble of $K$ interconnected OPFs, where each filter predicts plausible object tracks that are resilient to missing, noisy, and kinematically or dynamically infeasible measurements, thus bringing perceptional robustness. Through several interactive scenarios, we demonstrate that the proposed OPF approach provides robust tracking in human-robot interactive tasks agnostic to measurement type, even in the presence of prolonged and complete occlusion. Webpage: \url{https://opfilter.github.io/}. 
\end{abstract}
\section{Introduction}
\label{sec:intro}
As robots start escaping factories and coming into our homes and everyday lives, we must ensure that human-robot interaction (HRI) is safe, resilient, and robust. Much work in the HRI community focuses on developing safety-critical control schemes for interactive robots \cite{ROB-052, Li-RSS-21}. However, offering resilience and robustness relies not only on the strengths of controllers but also on the weaknesses of perception systems. Unreliable perception is the primary bottleneck for deploying HRI systems in the real world. 

In the context of multi-object multi-agent scenarios, the majority of modern tracking algorithms follow the `tracking-by-detection' paradigm, i.e., an object detector is used to identify objects in a camera frame, and the position and orientation of these objects are determined and linked into tracks. This approach heavily relies on the accuracy of object detection and motion prediction. However, in many HRI scenarios, the presence of occlusions is inevitable. Even the most advanced vision-based algorithms and hardware systems can fail in simple HRI tasks, such as tracking or handing over objects that are being manipulated or occluded by a human, another agent, or an object. Objects such as bikers or hand movements are inherently difficult to predict, and their trajectories are further complicated by the presence of partial or complete occlusions. Perception failures resulting from such occlusions can cause missing, noisy, or kinematically and dynamically infeasible measurements. If not addressed, these failures can negatively impact the expected behavior of the robot and threaten the safety of humans interacting with it. Rather than expecting a perfect perception system to arise, we instead assume that perception will always be unreliable in HRI systems and should be actively mitigated. In this work, we propose an approach inspired by human cognition and development to alleviate these inevitable, unreliable perception issues. According to Piaget's theory of cognitive development~\cite{Piaget}, humans develop an understanding of object permanence at an early age, i.e., they understand that objects continue to exist even when they are not visible or cannot be sensed. Object permanence is crucial to perception and memory and is a defining feature of human intelligence.

\begin{figure}[!tbp]
\begin{minipage}{0.425\linewidth}
  \centering
  \includegraphics[width=\textwidth]{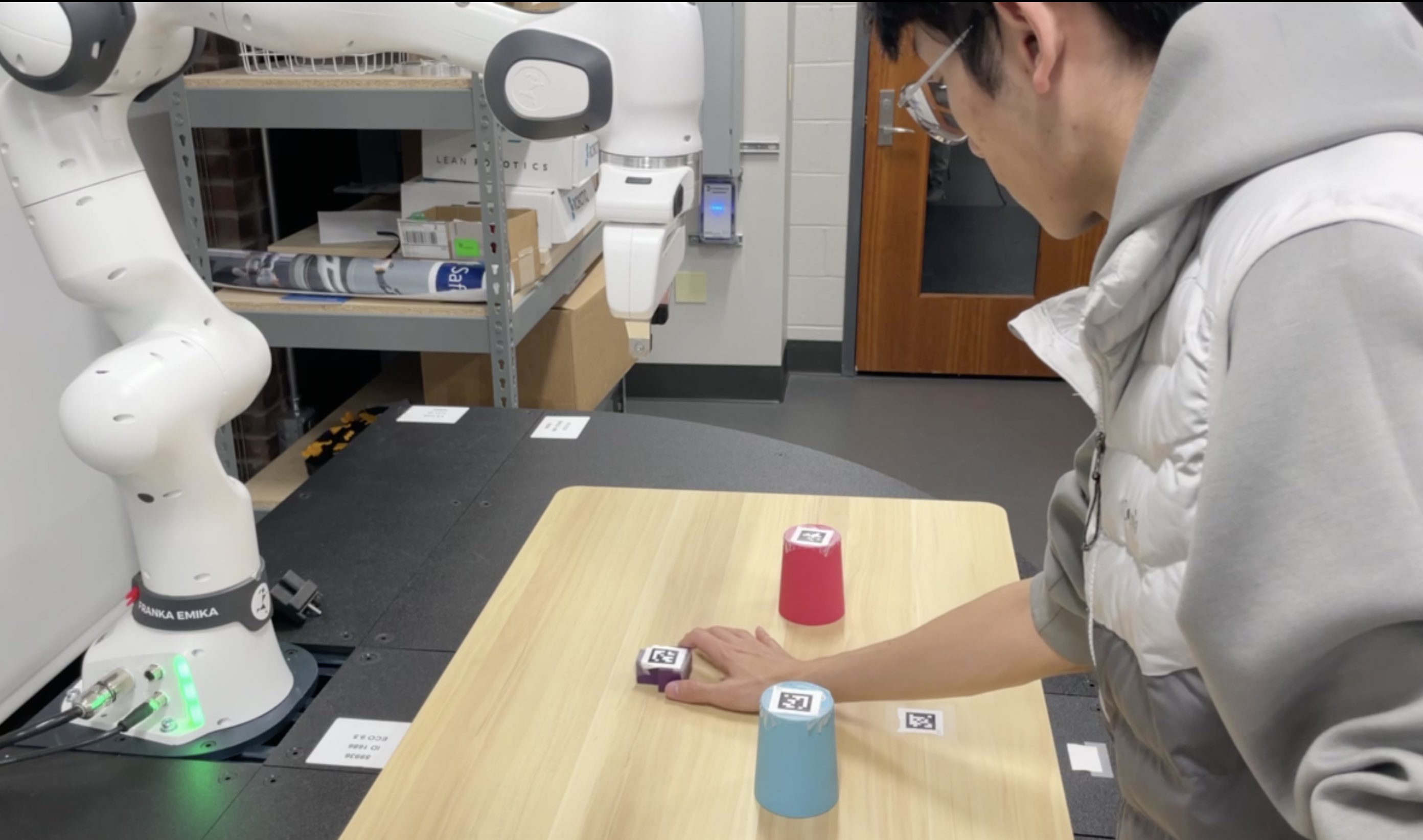}
  \includegraphics[width=\textwidth]{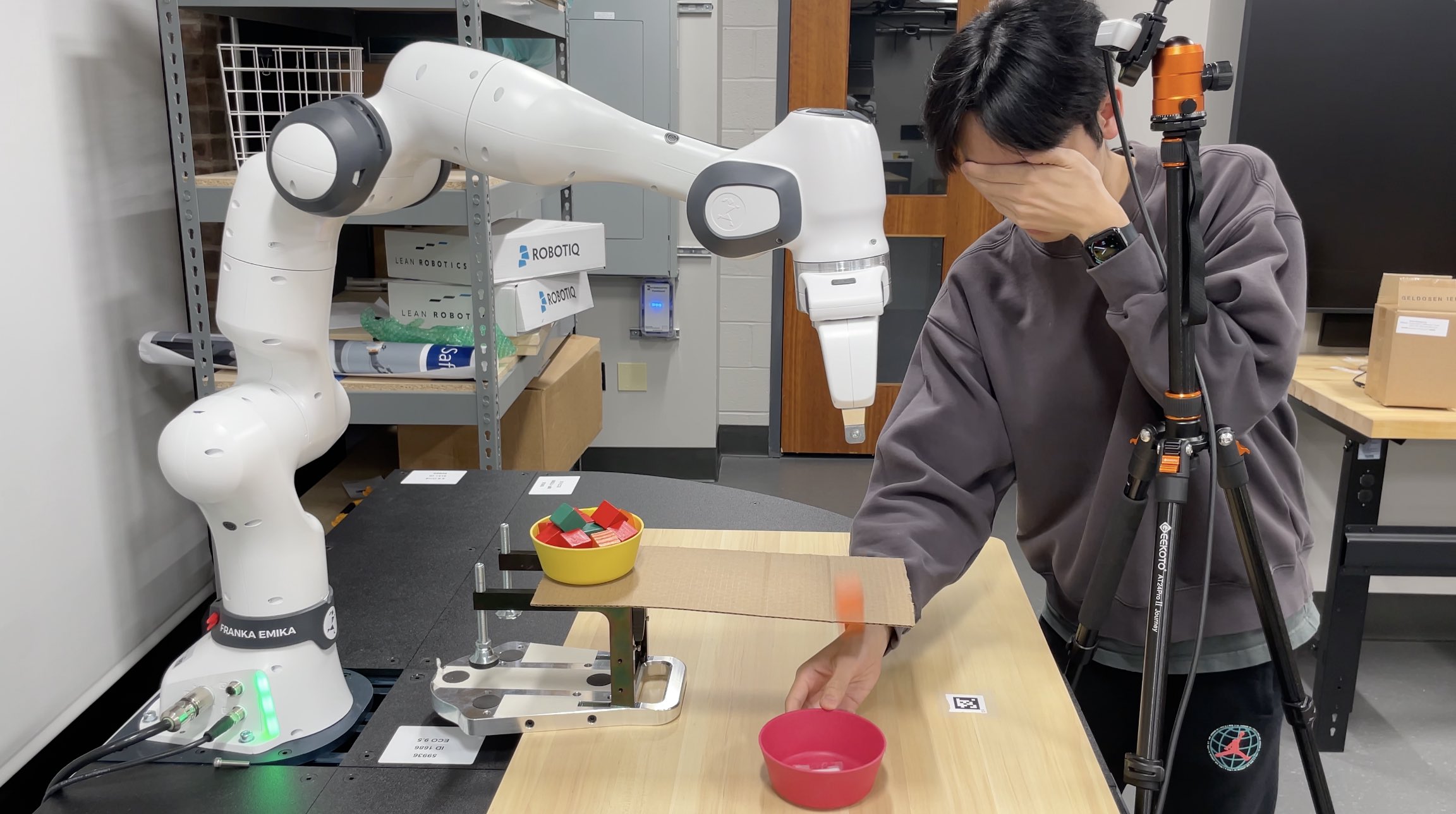}   
\end{minipage}\begin{minipage}{0.565\linewidth}
  \centering
  \includegraphics[width=\textwidth]{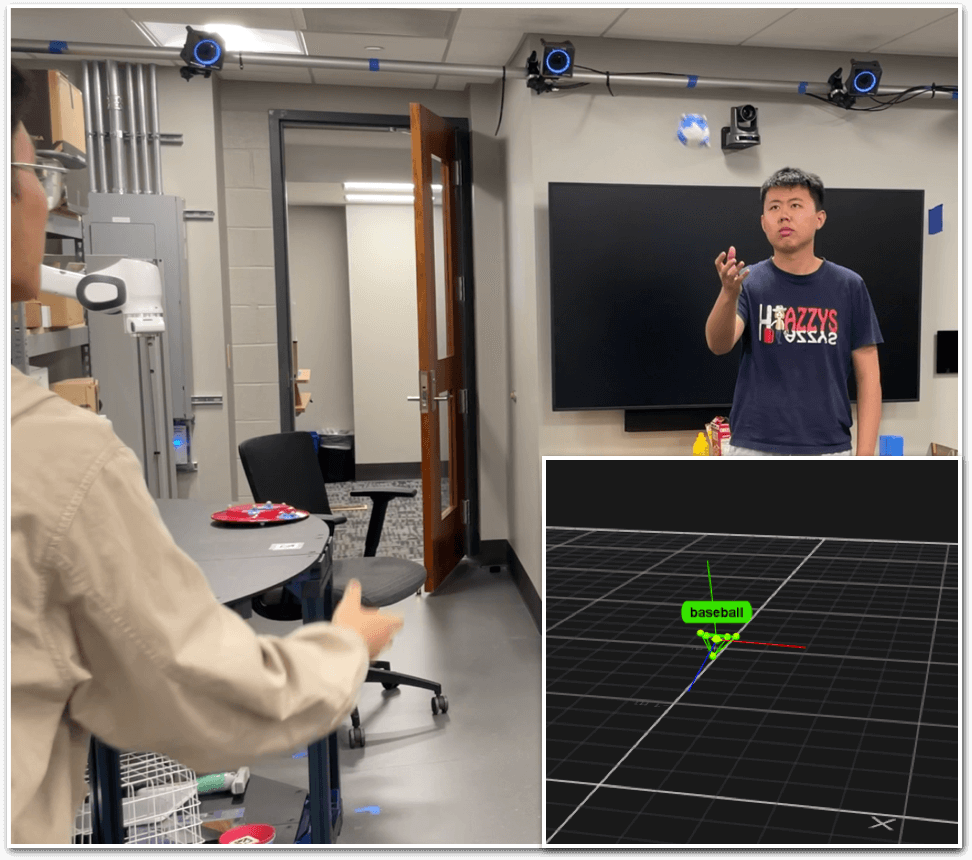}  
\end{minipage}
\vspace{-2.5pt}
\caption{\small Our Object Permanence Filter can be used for robust tracking of occluded objects or noisy measurements with different tracking systems at different frequencies. Apriltags (30Hz) used on the left for the cups game and sugar-dropping experiment and Optitrack (100Hz) used on the right for tracking a flying object. \label{fig:experiments}}  
\vspace{-20pt}
\end{figure}
Object permanence is evident in everyday life through various situations and experiences. A baby playing with a toy comprehends that the toy continues to exist even when it is concealed under a blanket; an adult who hides a car key in pockets understands that the key still exists. In addition, when a car drives behind a building, one can infer the car's location and expect it to reappear on the other side. Furthermore, object permanence plays a crucial role in more advanced forms of problem-solving, such as playing the cup game, during which an individual is aware of the ball being covered by one of the $N$ cups and can easily track the cup. Humans unequivocally use object permanence to make predictions about hidden objects and plan their actions accordingly. This capability to comprehend and reason about the persistence of objects is imperative for a broad range of activities and experiences. As demonstrated by Saiki \textit{et al.}~\cite{SAIKI2002133}, human beings are capable of maintaining multiple coherent object representations in dynamic scenarios, thereby enabling them to track multiple objects effectively. However, it becomes difficult as the number of objects increases. Hence, it is imperative for robots to develop their own robust models of object permanence. Developing such models can help robots understand the persistence of objects, leading to improved performance in various HRI tasks.

\textbf{Contributions:} We introduce the Object Permanence Filter (OPF) as a means to achieve resilient multi-object tracking. We modify the update step in particle filters with an Object Permanence Update (OP Update) that is robust to varying degrees of visual disruption. The OP update consists of three modules: \textit{dynamics, occluder} and \textit{uncertainty} modules, used to provide virtual measurements and scale covariance matrix when occlusions are detected (Fig. \ref{fig:OPF}). By incorporating object permanence rules into these modules, a robot's perception system can maintain track of objects even when they are partially or completely occluded. In addition, we introduce a \textit{feedback module} that monitors the uncertainty of the estimates and is used to modulate a robot's tracking behavior and inform the human operator if uncertainty explodes, allowing for a more safe, robust, and fluid HRI. We conduct comprehensive assessments of the OPF, employing both simulation and hardware experimentation. Our findings demonstrate the robust tracking capabilities of the OPF in heavy occlusion scenarios, alongside its capacity to adapt seamlessly to various measurement types.
\vspace{-4pt}
\section{Related Work}
\label{sec:rw}
\vspace{-1.5pt}
\textbf{6-DoF Object Tracking:} 6-DoF object tracking is an active area of research in computer vision (CV), encompassing the task of estimating the position and orientation of an object in 3D space. Various techniques have been proposed to address this problem, ranging from classical CV methods to learning-based approaches. As elucidated by Chen~\cite{Chen2012KalmanFF}, the Kalman filter has emerged as a prevalent technique in object tracking due to its ability to combine noisy measurements with a dynamic model of the object's position and orientation. In more recent developments, researchers are using deep neural networks to learn intricate, high-dimensional representations from data for better tracking performance~\cite{Labbe2020CosyPoseCM}. Nevertheless, the state-of-the-art approaches adhere to the tracking-by-detection paradigm, which can yield suboptimal results in occlusion scenarios. Though some work can alleviate it by memory mechanism~\cite{Wen2023BundleSDFN6}\cite{Hoorick2022RevealingOW} or sensor compensation~\cite{10160328}\cite{maskukf}, these methods primarily fall within the domain of learning-based approaches and may pose challenges when attempting to integrate them with other types of tracker and measurement. Besides, the rarity of depth and point cloud, as well as prolonged or extensive occlusion, invariably pose more challenges to the tracking performance.

\textbf{Occlusion handling:} Several research efforts have concentrated on addressing occlusions directly, including some modifications of the particle filter~\cite{Deng2019PoseRBPFAR}\cite{oapf}. For instance, Kourosh \textit{et al.} \cite{oapf} introduced an occlusion-aware particle filter tracker. During object occlusion, their approach resorts to a stochastic particle motion mechanism, resembling a random walk pattern, causing particles to disperse widely from the last known object position in a broader search endeavor. While they handle occlusions in a heuristic way, by doing random searches they do not integrate the concept of object permanence. In contrast, Tokmakov \textit{et al.} \cite{9710392} improved CenterTrack with a spatiotemporal recurrent neural network, introducing object permanence for joint detection and tracking. Their approach uses deterministic pseudo-ground-truth during occlusions by extrapolating object positions based on the last velocity. While similar in nature to the role of our \textit{dynamics module} described in Section \ref{sec:dynamics}, the concept of object permanence is not exploited elsewhere, including the more important aspect as introduced in Section \ref{sec:occluder}, and the NN-based tracker may diverge with wrongly generated pseudo-ground-truth labels, causing undesirable behaviors. Methods that consider object permanence thoroughly in the context of diverse occlusion scenarios are still in need.
\vspace{-2pt}
\section{OPF: Object Permanence Filter}
\label{sec:opfs}
\vspace{-0.5pt}
In this section, we introduce our proposed object permanence filter (OPF) framework. In section \ref{sec:filters}, we introduce the prediction and update equations for the particle filter as well as our usages, followed by the update rules to estimate the virtual observation $y_k^{\text{occ}}$ and covariance scaling function $\alpha_k$ for the PF updates in section \ref{sec:opf_update}. Finally, in Section \ref{sec:feedback}, we show that by monitoring the posterior covariance matrix, we can create cautious closed-loop tracking controllers and convey to humans that certain objects have been occluded for a long time to ensure safety.
\begin{figure*}[!tbp]
      \centering
      \includegraphics[width=0.9\textwidth]{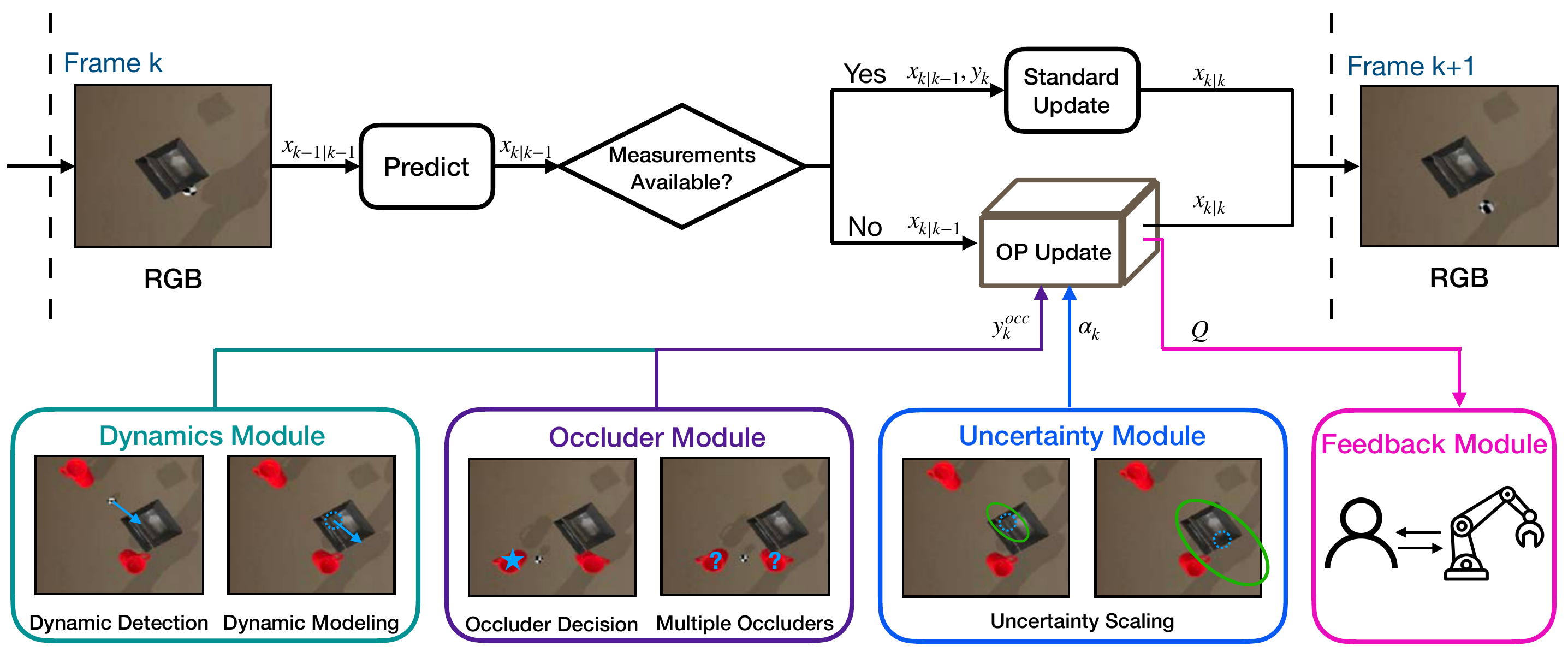}
      \caption{\small \textbf{Object Permanence Filter (OPF):} Following the predict-update cycle, the OPF introduces \textbf{OP update} when no measurements are available for the object being tracked. \textbf{OP update} consists of i) \textbf{dynamics module} to detect and model the dynamics of the object, ii) \textbf{occluder module} to help decide the occluder and deal with multiple occluders, and iii) \textbf{uncertainty module} to update the covariance matrix. The \textbf{feedback module} monitors the uncertainty of the updates by tracking the trace of the update covariance matrix, which can be used to change the behavior of the robot or indicate to the human operator when the uncertainty is above a safety threshold $\epsilon_{safe}$.}
      \label{fig:OPF}
      \vspace{-15pt}
  \end{figure*}
  \vspace{-1pt}
\subsection{PF for Object Tracking}
\label{sec:filters}
The particle filter works by generating and manipulating sets of particles to approximate the distribution of a stochastic process following a `predict + update' cycle.

\textbf{Importance Sampling} The key idea behind the PF is called importance sampling \cite{importance_sampling}, a technique that uses a known probability distribution $q(x)$ (proposal) to generate particles $x^{(i)}$ and approximate a given probability distribution $p(x)$ (target) by assigning weight to each of the particles:
\begin{align}
    &x^{(i)} \sim q, ~~ w^{(i)} = \frac{p(x^{(i)})}{q(x^{(i)})}, ~~ \forall i = 1,...,n
\end{align}

\textbf{Predict} Here we start with particles $x^{(i)}_{k|k}$:
\vspace{-5pt}
\begin{equation}
\begin{aligned}
P(x_k|y_1,...,y_k) = \frac{1}{n}\sum^n_{i=1}\delta_{x^{(i)}_{k|k}}(x)
\end{aligned} 
\end{equation}
where $\delta$ denotes the Dirac delta distribution, and all weights are equal, i.e. $w_{k|k}^{(i)} = \frac{1}{n}$. Suppose system dynamics $f$ with Gaussian noise $\epsilon_k$, then we can propagate each particle by one timestep $\forall i=1,\dots,n$:
\begin{align}
    &x^{(i)}_{k+1|k} = f(x^{(i)}_{k|k}, u_k) + \epsilon_k, ~\epsilon_k \sim N(0,R)
\end{align}
Particles weights are unchanged, i.e., $w_{k+1|k}^{(i)} = w_{k|k}^{(i)} = \frac{1}{n}$.

\textbf{Update} Then we update the weight of each particle using the likelihood of receiving a new observation $y_{k+1}$:
\begin{align}
\label{eq:pf_observation}
w^{(i)}_{k+1|k+1} = \eta P(y_{k+1}|x^{(i)}_{k+1|k})w^{(i)}_{k+1|k}
\end{align}
where $\eta$ is a normalization factor. 

Given a measurement function $g$, $P(y_{k+1}|x^{(i)}_{k+1|k})$ is a Gaussian with mean $g(x^{(i)}_{k+1|k})$,  variance $Q$, and depends on Gaussian observation noise $\nu_k$:
\begin{equation}
\label{eq:pf_update}
\resizebox{.85\hsize}{!}{$\begin{aligned}
    P(y_{k+1}|x^{(i)}_{k+1|k}) &= P\left(\nu_{k+1} \equiv y_{k+1} - g\left(x^{(i)}_{k+1|k}\right)\right)\\
    &= \frac{1}{\sqrt{(2\pi)^p \det(Q)}} \exp\left(-\frac{\nu^T_{k+1}Q^{-1}\nu_{k+1}}{2}\right)
\end{aligned}$}
\end{equation}

\textbf{Resampling} To avoid particle degeneracy problem~\cite{degeneracy}, resampling is proposed to remove unlikely particles with very low weights and effectively split the particles with very large weights into multiple particles:
\begin{align}
p(x) = \sum^n_{i=1} w^{(i)}\delta_{x^{(i)}}(x) \Rightarrow \frac{1}{n}\sum^n_{i=1}\delta_{x^{'(i)}}(x)
\end{align}

For our 6-Dof usage, an object's state is represented as 
\begin{equation}
x_k = (\xi_k^x, \xi_k^y, \xi_k^z, \theta_k, \phi_k, \psi_k) \in \mathbb{R}^6
\end{equation}
$(\xi_k^x, \xi_k^y, \xi_k^z)$ is the translation of the object in the current frame (in meters), and $(\theta, \phi, \psi)$ is the rotation, represented as Euler angles (in radians). Due to the high dimensionality and mismatch in the scales of translation and rotation, sampling the entire state can be inaccurate. Instead, we create two portions of particles to sample each, which improves sampling performance while maintaining decent time complexity.

\subsection{Object Permanence Update}
\label{sec:opf_update}
When a measurement from an object is missing due to occlusion or sensor failure at $t_{\text{occ}}$, then the state will remain the same for $k>t_{\text{occ}}$, which is $x_{k>t_\text{occ}|k>t_\text{occ}}\equiv x_{t_\text{occ}|t_\text{occ}}$. Due to object permanence, humans can understand that such missing measurements may be due to occlusions and can even predict the motion of that occluded object. To embed such a concept into the PF, in this work, we propose the Object Permanence (OP) Update that i) detects when occlusions happen, ii) infers the occluder from state estimates of neighboring objects, iii) estimates the occluded object dynamics, iv) updates state-estimate uncertainty.  

\textbf{OP Update Overview:} Let $\mathcal{O}_i$ be the $i$-th object being tracked in a set of $K$ objects/agents in the robot's workspace.

We start with the dynamics module (Section \ref{sec:dynamics}), which is used to detect whether object $\mathcal{O}_i$ was moving before occlusion. If so, we model the object dynamics to get future predictions as virtual measurements, which are continuously passed to the update stage until a new measurement appears.
If the object is static, we move to the occluder module (Section \ref{sec:occluder}), which is used to decide which object $\mathcal{O}_j,\forall j\neq i$ is the occluder by calculating and comparing the Bhattacharyya distances~\cite{bha}. The occluder's observation is used to update the measurement of the occluded object ${y_k^i}^{\text{occ}} \gets y_k^j$, depicted in Fig. \ref{fig:OPF} and listed in Alg. 1. This virtual observation ${y_k^i}^{\text{occ}}$ is fed to Eq. \ref{eq:pf_observation}. Finally, we scale the covariance matrix $Q$ in the update step of Eq. \ref{eq:pf_update} by a scalar function $\alpha_k^i$ (to increase uncertainty) as described in Section \ref{sec:op_uncertainty}.

\subsubsection{Dynamics Module}
\label{sec:dynamics}
Humans can not only realize the existence of an object that has been occluded via object permanence, but can also approximately guess the position of the occluded object~\cite{OP1}. In this work, we embed this property by maintaining a history of $H$ object states and analyzing the trajectory of the object before occlusion. Two trajectories of object $\mathcal{O}_i$'s translation $T$ and rotation $R$ before frame $k$ are defined as:
\begin{align}
\label{eq:H_track}
    ^{T}\text{Tr}^{\mathcal{O}_i}_{k-1} &= [(\xi_0^{x_i}, \xi_0^{y_i}, \xi_0^{z_i}), ..., (\xi_{k-1}^{x_i}, \xi_{k-1}^{y_i}, \xi_{k-1}^{z_i})]\\
    ^{R}\text{Tr}^{\mathcal{O}_i}_{k-1} &= [(\theta_0^i, \phi_0^i, \psi_0^i), ..., (\theta_{k-1}^i, \phi_{k-1}^i, \psi_{k-1}^i)]
\end{align}
If $\mathcal{O}_i$ is occluded at frame $k$, then we use the state of object $\mathcal{O}_i$ in past $H$ frames to decide for the dynamics of object $\mathcal{O}_i$. In this work, we set $H=50$:
\begin{itemize}[leftmargin=*]
    \item $^T\delta$ is a threshold to detect translation and $^R\delta$ is a threshold to detect rotation\footnote{For the current experiment setup, $^T\delta$ is set to 0.01 and $^R\delta$ is set to 0.5}, an object is regarded as \textbf{static} if,
    \begin{equation}
    \begin{aligned}
    &\max(\|{^T\text{Tr}^{\mathcal{O}_i}_{k-1}[p]} - {^T\text{Tr}^{\mathcal{O}_i}_{k-1}[q]}\|_2) \leq {^T\delta}  ~~ \text{AND} \\
    &\max(\|{^R\text{Tr}^{\mathcal{O}_i}_{k-1}[p]} - {^R\text{Tr}^{\mathcal{O}_i}_{k-1}[q]}\|_2) \leq {^R\delta},
    \\
    &\forall p, q \in \{k-H,...,k-1\}
    \end{aligned}
    \end{equation}
    Then we move to occluder module (Section \ref{sec:occluder}) and use the detected occluder's $\mathcal{\bar{O}}_j, \forall j\neq i$ observation $y_k^j$ as the object $\mathcal{O}_i$'s virtual measurement ${y_k^i}^{\text{occ}}$.
    \item Correspondingly, an object is regarded as \textbf{moving} if,
    \begin{equation}
    \begin{aligned}
    &\max(\|{^T\text{Tr}^{\mathcal{O}_i}_{k-1}[p]} - {^T\text{Tr}^{\mathcal{O}_i}_{k-1}[q]}\|_2) > {^T\delta} ~~ \text{OR} \\
    &\max(\|{^R\text{Tr}^{\mathcal{O}_i}_{k-1}[p]} - {^R\text{Tr}^{\mathcal{O}_i}_{k-1}[q]}\|_2) > {^R\delta},\\
    &\exists p, q \in \{k-H,...,k-1\}
    \end{aligned}
    \end{equation}
    Then we predict the next state of object $\mathcal{O}_i$ by fitting a first-order polynomial $\text{poly}(t) = at + b$ with $a,b \in \mathbb{R}^4$ for translation $T$ and rotation $R$, which is determined by minimizing the squared error:
    \begin{align}
    \label{eq:polyfit}
        & {a}, {b} = \argmin_{a,b} \sum^{k-1}_{p=k-H}\left|\text{poly}(p)-Y_p^i\right|^2
    \end{align}
    Here $Y_p^i \in \mathbb{R}^4$ consists of a three-dimensional translation vector $(\xi_p^{x,i}, \xi_p^{y,i}, \xi_p^{z,i})$ and one rotation scalar $\omega^i$. To get $\omega^i$ from $(\theta^i, \phi^i, \psi^i)$, a transformation is executed whereby the Euler angles are converted into an axis-angle representation $([e_x^i, e_y^i, e_z^i], \omega^i)$ with the primary objectives of enhancing tracking efficiency by only fitting one variable $\omega^i$ around the fixed axis, and mitigating the computational complexity associated with the fitting of all three Euler angles. This transformation serves to avoid the sine waves of Euler angles which are less efficient to fit. The outcome of this fitting is well-predicted rotation and the achievement of real-time (100Hz) object permanence tracking. Once $a,b$ are determined, we don't update them by fitting a new model using the newly predicted positions to avoid the accumulation of errors. Dynamic predictions stop once the measurement of this object appears again. 
\end{itemize}
\subsubsection{Occluder Module}
\label{sec:occluder}
We introduce the Bhattacharyya distance~\cite{bha}, which measures the similarity of two objects $\mathcal{O}_p, \mathcal{O}_q$'s probability distributions $P,Q \in \chi$:
\begin{align}
\label{eq:B-distance}
    &D_B(\mathcal{O}_p,\mathcal{O}_q) = -\ln\left(BC(P,Q)\right)\\
    &BC(P,Q) = \int_{\chi} \sqrt{p(x)q(x)}dx \notag
\end{align}
If $\mathcal{O}_i$ is occluded, Eq. \ref{eq:B-distance} between $\mathcal{O}_i$ and all other $K-1$ objects is used to obtain two potential occluders $\mathcal{O}_p,\mathcal{O}_q$ with smallest distances $(D_B(\mathcal{O}_i, \mathcal{O}_p) < D_B(\mathcal{O}_i, \mathcal{O}_q))$. A hyperparameter $\epsilon_{\text{occ}} = 0.01$ based on the size of the tested object is set to distinguish single/multiple occluder:
\begin{itemize}[leftmargin=*]
    \item Single Occluder: if $D_B(\mathcal{O}_i, \mathcal{O}_q)-D_B(\mathcal{O}_i, \mathcal{O}_p) > \epsilon_{\text{occ}}$, then $\mathcal{O}_p$ is regarded as occluder and provides virtual observation ${y_k^{i}}^{\text{occ}} \gets y_k^p $.
    \item Multiple Occluders: if $D_B(\mathcal{O}_i, \mathcal{O}_q)-D_B(\mathcal{O}_i, \mathcal{O}_p) \leq \epsilon_{\text{occ}}$, then a virtual object $\mathcal{O}_{i^{'}}$ with state $x_k^{i^{'}}$ copied from $x_k^i$ is created, and two virtual observations are created: ${y_k^{i}}^{\text{occ}} \gets y_k^p, {y_k^{i^{'}}}^{\text{occ}} \gets y_k^q  $. Both $\mathcal{O}_{i}$ and $\mathcal{O}_{i^{'}}$ will continue for the `predict-update' cycle until the observation for it appears again. Hence, uncertainty is introduced by two sets of states, and the robot is aware of the multiple occluders.
\end{itemize}
We do not pre-set the objects as occluders (e.g. hands or end-effector) or occluded objects, because sometimes so-called `occluders' can also be occluded by objects or other occluders. In this work, we treat every tracked object equally. 
\newcommand{\nosemic}{\renewcommand{\@endalgocfline}{\relax}}
\newcommand{\dosemic}{\renewcommand{\@endalgocfline}{\algocf@endline}}
\newcommand{\pushline}{\Indp}
\newcommand{\popline}{\Indm\dosemic}
\let\oldnl\nl
\newcommand{\nonl}{\renewcommand{\nl}{\let\nl\oldnl}}
\newcommand\mycommfont[1]{\footnotesize\ttfamily\textcolor{blue}{#1}}
\SetCommentSty{mycommfont}
\SetKwInOut{Input}{Input}
\SetKwInOut{Output}{Output\,}
\SetKwInOut{Parameters}{Parameters}
\SetKwProg{Tree}{Tree}{}{EndTree}
\setlength{\textfloatsep}{0pt}
\begin{algorithm}[!tbp]
\label{alg:op_updates}
    \small{
    \caption{Object Permanence Update Step for $\mathcal{O}_i$}
    \Input{Past $H$ trajectory $^{T, R}Tr_{k-1}^{\mathcal{O}_i}$ (Eq. \ref{eq:H_track}) of $\mathcal{O}_i$\\
    Current state $x_{k|k-1}$ of object $\mathcal{O}_i$\\
    Current state $x_k^j$ and measurements $y_k^j$ \\ of all objects
    $\mathcal{O}_j~\forall j=1,\dots,K \setminus i$\\
    }
    \Output{Virtual observation ${y_{k}^i}^{\text{occ}}$ and covariance matrix scaling factor $\alpha_k^i$ for $\mathcal{O}_i$
    }
    \Parameters{$H,\kappa,\epsilon_{\text{occ}}$}

    \tcc{Dynamics Module}
    \For{$p \in [k-H,k-1]$}{
    \For{$q \in [k-H,k-1]$}{
    \If{$(\|{^T\text{Tr}}^{\mathcal{O}_i}_{k-1}[p]-{^T\text{Tr}}^{\mathcal{O}_i}_{k-1}[q]\|_2 > 0.01$ \textbf{or} $\|{^R\text{Tr}}^{\mathcal{O}_i}_{k-1}[p]-{^R\text{Tr}}^{\mathcal{O}_i}_{k-1}[q]\|_2 > 0.01)$}{
    \tcc{Fit the polynomial}
    ${y_{k}^i}^{\text{occ}} \gets \text{poly}(x)$ \text{ with Eq.} \ref{eq:polyfit}\\
    $\alpha_k^i \gets \text{Compute from}~ x_k^p~\text{with Eq.}$~\ref{eq:alpha_scale}\\   
    \Return (${y_{k}^i}^{\text{occ}},\alpha_k^i$)
    }
    }
    }
    
    \tcc{Occluder Module}
    $\text{Dists} \gets $ []\\
    \For{$j \in [1,K]\setminus i$}{
    $D_B(\mathcal{O}_i, \mathcal{O}_j) \gets \text{Compute with Eq.}$ \ref{eq:B-distance}\\
    AddItem($\text{Dists}, D_B(\mathcal{O}_i, \mathcal{O}_j)$)
    }
    $\text{sort}(\text{Dists})$ ascendingly\\
    $D_B(\mathcal{O}_i, \mathcal{O}_p),D_B(\mathcal{O}_i, \mathcal{O}_q) \gets \text{Dists}[0], \text{Dists}[1]$ \\
    \If{$D_B(\mathcal{O}_i, \mathcal{O}_q)-D_B(\mathcal{O}_i, \mathcal{O}_p) > \epsilon_{\text{occ}}$}{${y_{k}^i}^{\text{occ}} \gets y_k^p$\\
    $\alpha_k^i \gets \text{Compute with Eq.}$~\ref{eq:alpha_scale}\\
    \Return (${y_{k}^i}^{\text{occ}},\alpha_k^i$)}
    
    \Else{${y_{k}^i}^{\text{occ}}, {y_{k}^{i'}}^{\text{occ}} \gets y_k^p,y_k^q$\\
    $\alpha_k^i, \alpha_k^{i'} \gets \text{Compute from}~ x_k^p,x_k^q~\text{with Eq.}$~\ref{eq:alpha_scale}\\
    \Return (${y_{k}^i}^{\text{occ}},\alpha_k^i$, ${y_{k}^{i'}}^{\text{occ}},\alpha_k^{i'}$)}}
\end{algorithm}

\vspace{-10pt}
\subsubsection{Uncertainty Module}
\label{sec:op_uncertainty}
In the PF, one can measure the uncertainty of the filter through its covariance matrices. In the non-occluded condition, the covariance matrix updates at each step according to Eq. \ref{eq:pf_update}. However, when the object is occluded we use $y_k^i \gets y_k^j$ with $y_k^j$ being the observation of the occluder $\mathcal{O}_j$. Hence, we artificially increase the corresponding covariance matrices for each time step that measurement is missing as $\tilde{Q} \gets \alpha_k Q$ in Eq.~\ref{eq:pf_update} with: 
\begin{equation}
\label{eq:alpha_scale}
\alpha_k = \kappa^v
\end{equation}
an exponentially increasing function dependent on the velocity $v$ calculated from the object's trajectory and constant $\kappa > 1$. We empirically found $\kappa=1.03$ suitable to showcase a moderate exponential increase in uncertainty.
\vspace{-2.5pt}
\subsection{Feedback Module}
\label{sec:feedback}
\vspace{-2.5pt}
\subsubsection{Human Intervention Feedback} Notice that when an object is occluded for a long time the scaled covariance matrices $\tilde{Q}$ for the PF will explode. This is by design as we can indicate the uncertainty as:
\begin{equation}
\label{eq:uncertainty_trace}
\mathcal{U}_{\text{PF}} = \text{trace}(\tilde{Q}) 
\end{equation}
Hence, given an uncertainty safety threshold $\epsilon_{\text{safe}}$ if $\mathcal{U}_{PF}\geq\epsilon_{\text{safe}}$ the robot should either stop, fall back to a safety mode or send an alert signal to the human operator. 
\begin{figure*}[!tbp]
\centering
\begin{minipage}{0.21\textwidth}
  \centering
  \includegraphics[width=\linewidth]{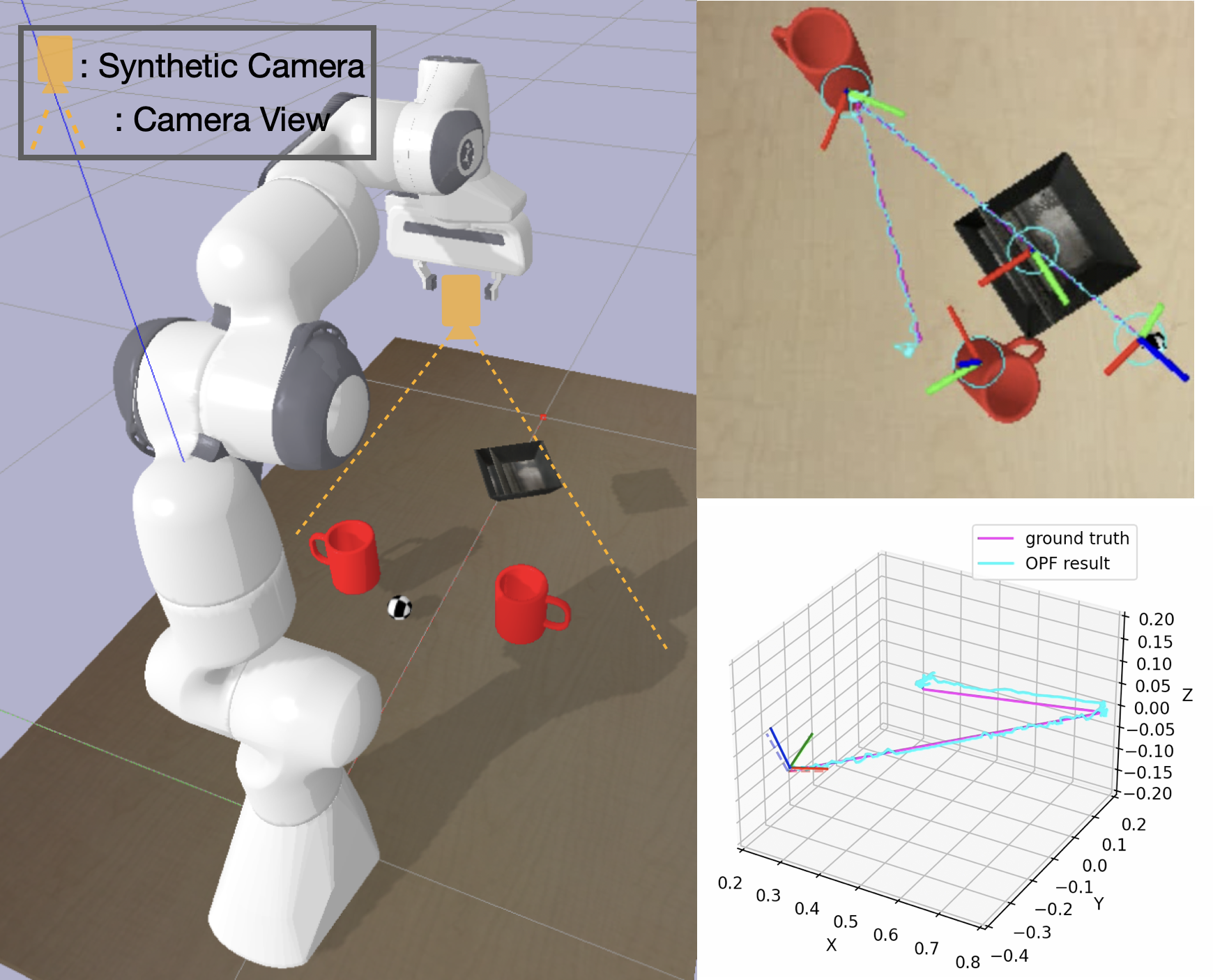}
  \vspace{4pt}
  \includegraphics[width=\linewidth]{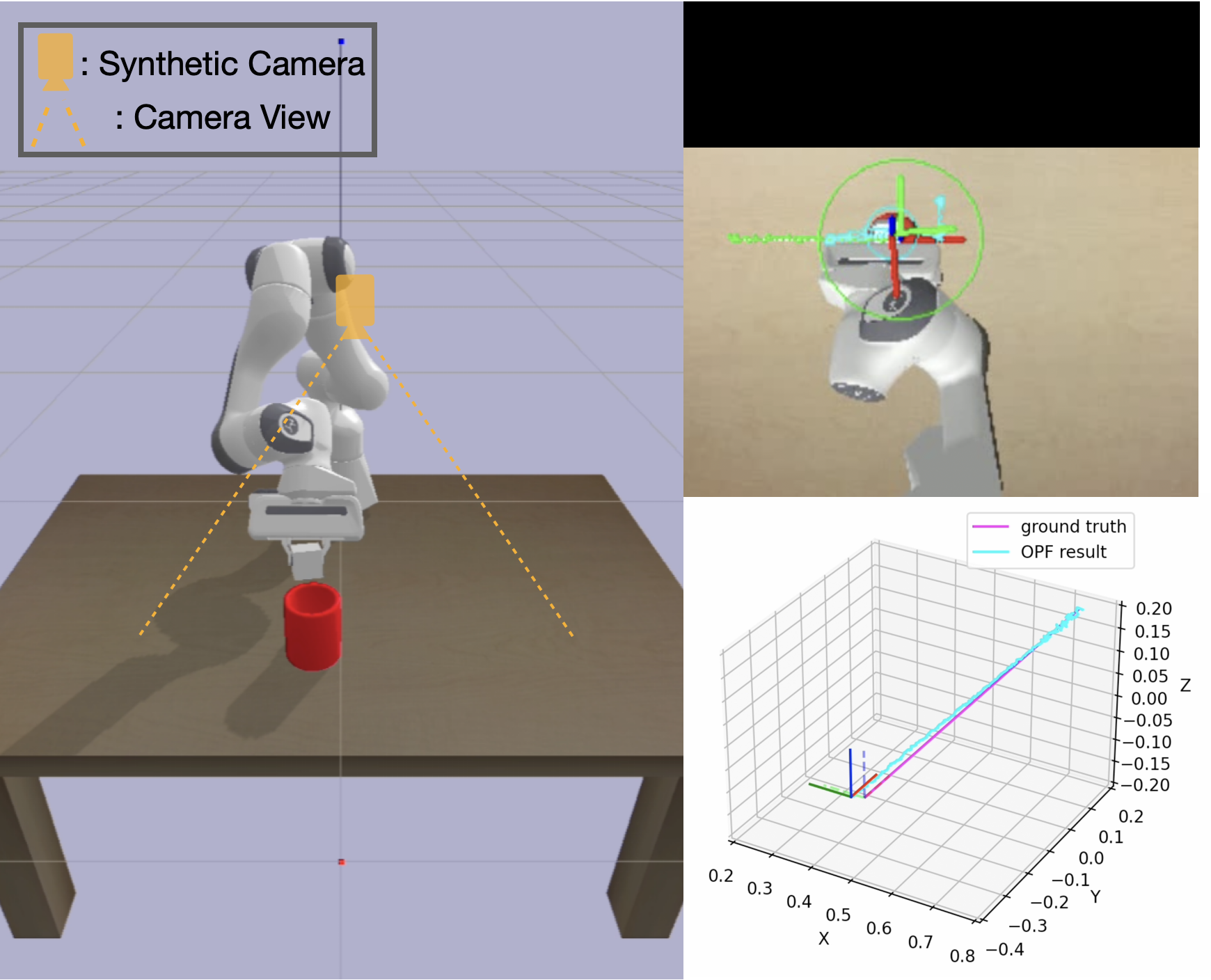}
\end{minipage}
\begin{minipage}{0.715\textwidth}
\includegraphics[width=\linewidth]{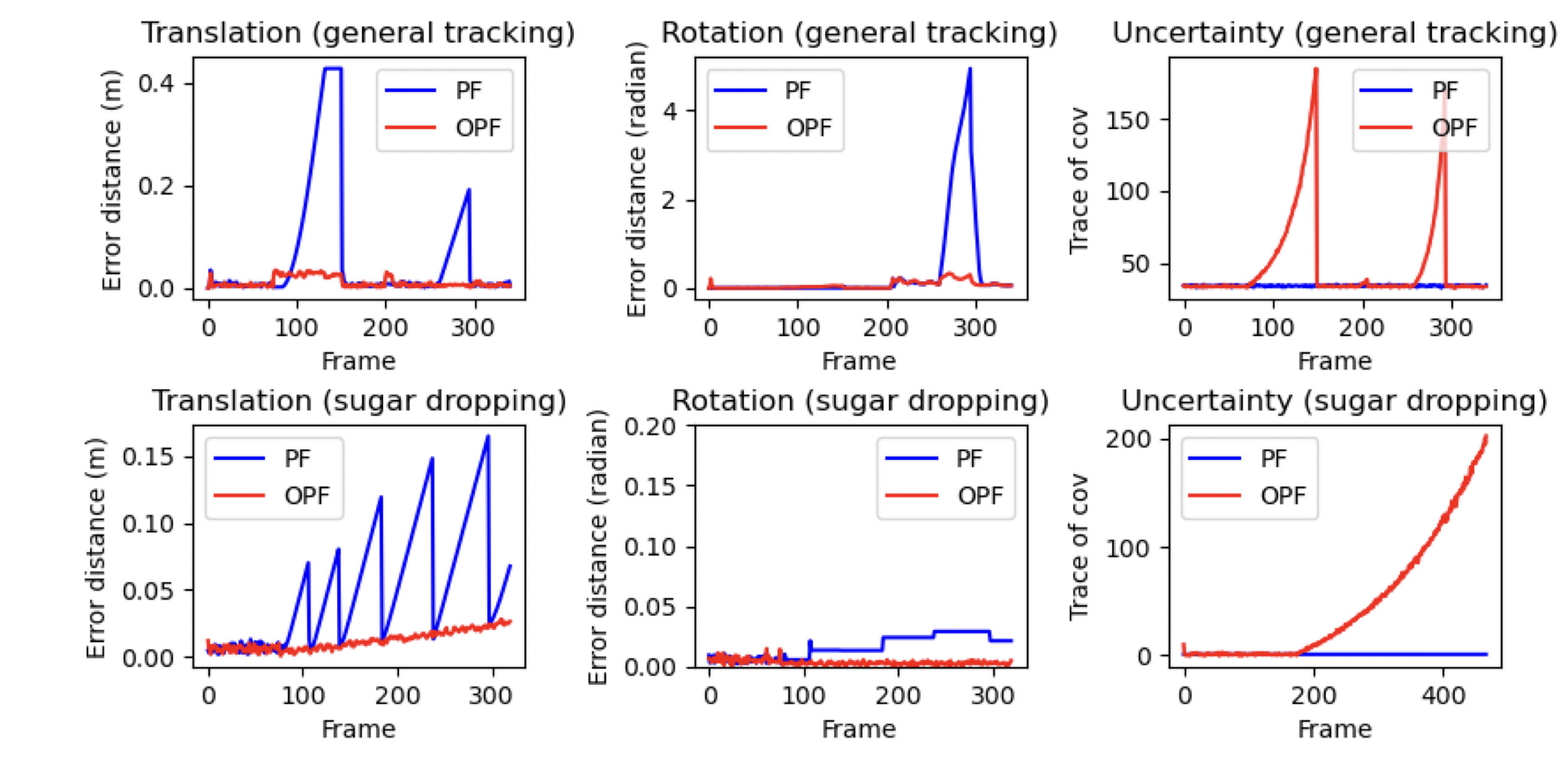}
\end{minipage}
\caption{\small \textbf{Comparative Results:} (top row) General object permanence (OP) tracking experiment, (bottom row) sugar-dropping experiment (inspired by \cite{maskukf}). The X-axis for all plots denotes $k$-th camera frame. (1st column) simulation snapshots, (2nd-3rd columns) tracking error distances of translation and rotation of the occluded object given by \textcolor{blue}{PF (blue)} and \textcolor{red}{OPF (red)} (4th column) traces of $Q$ (Eq. \ref{eq:pf_update}).}
\label{fig:results}
\vspace{-5pt}
\end{figure*}
\begin{table*}[t]
\begin{center}
\centering
\begin{tabular}{ |P{2.9cm}|P{2.55cm}|P{2.55cm}|P{2.55cm}|P{2.55cm}|  }
 \hline
  & \multicolumn{2}{c|}{\textbf{General OP Tracking}} & \multicolumn{2}{c|}{\textbf{Sugar-dropping}} \\
 \hline
 \textbf{Tracking Error} & \textbf{With OPF} & With Standard PF & \textbf{With OPF} & With Standard PF\\
 \hline
 Translation error & \cellcolor{blue!15} 0.01138 $\pm$ 7.253e-4 & 0.06289 $\pm$ 6.498e-4 & \cellcolor{blue!15} 0.02129 $\pm$ 5.872e-4 & 0.05018 $\pm$ 6.936e-4\\
 \hline
Rotation error & \cellcolor{blue!15} 0.06869 $\pm$ 2.947e-3 & 0.3870 $\pm$ 1.270e-3 & \cellcolor{blue!15} 0.003411 $\pm$ 4.629e-4 & 0.01456 $\pm$ 2.274e-3 \\
 \hline
\end{tabular}
\caption{\small Numerical results for general OP tracking (experiment 1) and sugar-dropping (experiment 2) over 5 runs.}\label{tab:results}
\label{tab}
\end{center}
\vspace{-20pt}
\end{table*}

\subsubsection{Closed-loop Tracking Controller} Another use of $\mathcal{U}_{PF}$ defined in Eq.~\ref{eq:uncertainty_trace} can be to create a cautious tracking controller when used in closed-loop with the output of the filter. Let $\xi_r,\dot{\xi}_r,\xi_o, \dot{\xi}_o\in\mathbb{R}^3$ be positions and velocities of the robot's end-effector and object to track, respectively. Then one can define an object tracking control law as follows \cite{thebook}:
\begin{equation}
\label{eq:controller_tracking}
    \dot{\xi}_r = - k_p(\mathcal{U}_{PF})\left(\xi_r - \xi_o\right) +  k_d(\mathcal{U}_{PF})\dot{\xi}_o
\end{equation}
where $k_p(\mathcal{U}_{PF}) \in [0,k_p^{\text{nom}}]$ is a positive bounded tracking gain and $k_d(\mathcal{U}_{PF})\in [0,1]$ a feedforward velocity damping term, formulated as decreasing sigmoid functions $
    k_{(\cdot)}(\mathcal{U}_{PF}) = k^{\text{nom}}_{(\cdot)}\cdot\frac{(\frac{1}{2}\epsilon_{\text{safe}})^n}{(\frac{1}{2}\epsilon_{\text{safe}})^n + \mathcal{U}_{PF}^n}$
with $n\geq 1$ controlling the steepness of the transition of tracking gain $k^{\text{nom}}_{(\cdot)} \rightarrow 0$ as $\mathcal{U}_{PF}$ changes from $0\rightarrow \epsilon_{\text{safe}}$. Thus, nominal gains $(k^{\text{nom}}_{(\cdot)}, d^{\text{nom}}_{(\cdot)})$ are used when $\mathcal{U}_{PF}=0$ and decrease to 0 as $\mathcal{U}_{PF}\rightarrow \epsilon_{\text{safe}}$.
\vspace{-2.5pt}
\section{Experiments and Results}
\label{sec:exp}
\vspace{-2.5pt}
In this section, we first introduce our evaluation metrics, followed by tracking performance compared to the ground truth and the standard PF in simulated experiments (Fig. \ref{fig:results}), as well as interesting hardware experiments designed to showcase the strength of our OPF (Fig. \ref{fig:experiments}). Videos are provided in the multimedia attachment and project webpage.
\vspace{-15pt}
\subsection{Metrics and Evaluation Protocol}
\vspace{-1.5pt}
Following we describe the two metrics we use to evaluate the the PF and OPF on two occlusion-heavy simulated tasks: 
\begin{itemize}[leftmargin=*]
    \item \textit{Error distance:} Error distances for translation and rotation are defined by $\frac{\sum_{k=0}^n \|\hat{p}_k-p_k\|_2}{n}$, with $p_k = (\xi_k^x, \xi_k^y, \xi_k^z)$ for translation and $p_k = (\theta_k, \phi_k, \psi_k)$ for rotation of a tracked object at the $k$-th frame.
    \item \textit{Confidence:} Another measure of the effectiveness of the OPF is the amount of confidence for each prediction. Following the feedback module (Section \ref{sec:feedback}) we evaluate confidence for each filter proportional to the uncertainty; i.e., $\text{trace}(\cdot)$ of covariance matrices via Eq. \ref{eq:uncertainty_trace}.
\vspace{-5pt}    
\end{itemize}
\vspace{-5pt} 
\subsection{Comparative Simulation Experiments}
\label{sec:comparative_sim}
\vspace{-2.5pt}
We present two evaluation experiments implemented in PyBullet, (experiment 1) general OP tracking with both static and moving objects and (experiment 2) self-occluding sugar-dropping task as in \cite{maskukf}. Table \ref{tab:results} shows evaluation metrics on both tasks compared to the standard PF for 5 runs. Fig. \ref{fig:results} shows plots of the objects of interest predicted position vs. ground truth position and covariance traces w.r.t. time. 

In all simulations, the origin of the coordinate system is placed on the base frame of the Franka Research 3 arm. We assume a fixed external camera, see Fig. \ref{fig:results}. All objects are tracked via color-based segmentation and blob detection on the RGB images from the synthetic camera. We use 5000 particles for translation and rotation per object for the PF. 
\subsubsection{General OP Tracking} This experiment is intended to showcase the performance of the standard PF vs. OPF while tracking multiple self-occluding dynamic and static objects.

\textbf{Experiment Details:}  The object of interest in this experiment is a ball set initially at $[0.4,0,0]$. Two dynamic mugs (denoted as left/right wrt. the camera view) are set at $[0.4, \pm 0.15, 0.03]$. A static tray is at $[0.55,-0.09,0.2]$. The camera is mounted above the table at $[0.55,0,0.49]$ and looks downward. We divide this experiment into two steps to test static objects and moving objects under occlusion scenarios:

\underline{Step one:} The two mugs will move to the center of the table and cover the ball. The ball is closer to the left mug: $[0.4,0.06,0]$. Then the left mug is moved to $[0.8,0.15,0.03]$ and the ball moves with it. After the ball gets to the target position it is revealed and then occluded again with the same mug. This step tests the OPF tracking of a static object being occluded by multiple possible dynamic occluders.

\underline{Step two:} The ball starts moving to $[0.3, -0.35, 0]$, leading it to pass under the tray. This step is designed to test the OPF with a moving object.

\textbf{Analysis and Results} The results are shown in Fig. \ref{fig:results} (top row) and Table \ref{tab}. \underline{Step 1} starts at frame 90, and \underline{Step 2} starts at frame 210. When using the standard PF one can see an obvious error distance in tracking. Such error arises from missing measurements, the PF will not propagate the state ahead, so the ball's predicted state is always close to where it was occluded, and abruptly updates to the position where it is revealed. The OPF on the other hand, is capable of predicting the motion of the occluded object very accurately (as shown in the plots and tracking error statistics), with an exponential increase in uncertainty by design (as shown in the trace plot). Further, during frames 250-300 in \underline{Step 2}, the ball is moving under the tray. Without the OP update, the standard PF fails to predict the ball's position under the tray and performs a sudden update from the top to the bottom of the tray, while the OPF provides a much smoother approximation. 

Regarding \textit{uncertainty}, one can see in the last column of Fig. \ref{fig:results} the trace of the covariance matrix exhibiting 2 peaks increasing exponentially as the ball is occluded two times. For the first one it is occluded passively by the left mug, and the second one is caused by actively moving below the tray. The peaks exhibit faster and larger increases as they depend on the velocity and the timesteps that the object is occluded as per Eq. \ref{eq:alpha_scale}. When the measurement of the ball occurs again, the covariance drops to a normal scale. 

\subsubsection{Sugar-dropping Experiment}
We showcase a dynamic sugar-dropping task, where a robot is tracking a mug with an external camera positioned prone to self-occlusions by the end-effector (inspired by \cite{maskukf}). The robot tracks the predicted state of the moving mug with a tracking controller as Eq. \ref{eq:controller_tracking}.

\textbf{Experiment Details} The robot grasps the sugar cube and then the mug is moved from $[0.55, 0.45, 0]$ to $[0.55, -0.45, 0]$. The robot will start tracking the mug as it moves to $[0.55, 0.25, 0]$ until the end. The camera is mounted at $[0.55, 0, 0.8]$ and looks downward. This experiment emulates a robot trying to drop a sugar cube into a cup of moving coffee, but the coffee cup is occluded by its own hand.

\textbf{Analysis and Results} In this case, the mug is occluded by the robot itself. With the OP rule, the robot is aware of the existence of the mug's dynamics and tries to approximate it, providing a smooth tracking trajectory under occlusion. In comparison, the performance of this task is erratic using the standard PF. The robot will stay still when self-occlusion exists, and sharply update to the newly observed state, resulting in a self-occlusion again and undesired jerky motion. As shown in Fig. \ref{fig:results} and Table \ref{tab}, the standard PF is laggy and jerky, whereas the OPF variants approximate the dynamics of the mug very close to the ground truth trajectory. Further, one can see the exponentially growing trace of the covariance matrix during the long self-occlusion conveying uncertainty. 
\subsection{Hardware Experiments}
\vspace{-1pt}
In this section, we conduct OPF-based tracking experiments with Franka Research 3 and test its generalizability with different measurement types. While these experiments have no ground truth they show the strength of the proposed OPF in different applications and scenarios.
\subsubsection{Cups Game (Hardware)} We validate the OPF on the cups game as shown in Fig. \ref{fig:experiments}. We track the 2 occluding cups and the object of interest with AprilTags and human hands with Mediapipe~\cite{mp}. Three users are invited to freely move around any of the objects with different occluding sequences and moving direction/speed as in the original cups game. Results show the robot manipulator equipped with OPF can robustly track the object under occlusion of different kinds of occluders, and the robot is moving smoothly without jerky motions caused by sudden updates in the standard PF. Due to the feedback module, the robot is able to stop actively because of exploding uncertainty.
\subsubsection{Sugar-dropping (Hardware)} We validate the OPF on a variant of the sugar-dropping experiment from Section \ref{sec:comparative_sim}. To emulate self-occlusions we include a large static occluder that hides the bowl when it is moving. The robot manipulator tracks the bowl and releases the sugar as soon as its confidence of the bowl's state is high (low uncertainty) after occlusion. Three blindfolded users are invited to generate different sugar-dropping trajectories and initial conditions. Results show the robot dropping the sugar in the bowl accurately with a high probability $8/10$. 
\subsubsection{Different Measurement Types} Finally, we validate that the OPF can be built on different measurement systems:
\begin{enumerate}
    \item \textbf{CV-based pose tracker}: We conduct experiments on the YCB dataset\cite{alli2017YaleCMUBerkeleyDF} by putting the mustard bottle into a cookie box completely utilizing BundleSDF \cite{Wen2023BundleSDFN6}, showcasing OPF's ability to handle complete and prolonged occlusion.
    \item \textbf{Marker-based motion capture system}: We conduct a baseball-throwing experiment, during which the motion capture system is disabled for 0.5 seconds. The OPF succeeds in modeling the parabola of baseball during that period and provides robust 6-DoF tracking.
\end{enumerate}

\centerline{More details can be found on our project website:} 
\centerline{\url{https://opfilter.github.io/}}

\section{Conclusion and Future Work}
\vspace{-1pt}
In this work, we propose a set of assumptions and rules to computationally embed object permanence into the particle filter to form the object permanence filter (OPF), which is an extended 6-DoF filter robust to heavy and prolonged occlusion scenarios in interactive tasks providing plausible tracking. We show that the OPF works well in simulation and hardware experiments, and due to its agnostic nature can be easily applied to multiple existing 6-DoF trackers achieving real-time performance. We are currently working on extending this framework to consider physics-based object dynamics as well as kinematic constraints for occlusion-aware human skeleton tracking and handover scenarios. Finally, an automated tuning of parameters like $\delta$ and $\epsilon_{\text{occ}}$ can be dependent on the actual 3D shape of the object to get more generalizable thresholds for moving and occlusion.

\addtolength{\textheight}{-12cm}   

\bibliographystyle{IEEEtran}
\bibliography{citations.bib}

\end{document}